\documentclass[runningheads]{llncs}
\usepackage[T1]{fontenc}
\usepackage{graphicx}
\usepackage{accv}
\usepackage{accvabbrv}
\usepackage{booktabs}
\usepackage[accsupp]{axessibility}
\usepackage{wrapfig}
\usepackage{hyperref}
\usepackage{color}

\makeatletter
\def\@seccntformat#1{\@ifundefined{#1@cntformat}%
   {\csname the#1\endcsname\quad}  
   {\csname #1@cntformat\endcsname}
}
\let\oldappendix\appendix 
\renewcommand\appendix{%
    \oldappendix
    \newcommand{\section@cntformat}{\appendixname~\thesection\quad}
}
\makeatother
\begin{document}
\title{Unsupervised Video Summarization via Iterative Training and Simplified GAN}
\titlerunning{Unsupervised Video Summarization}
%
\author{Hanqing Li\inst{1}\orcidID{0009-0003-5354-7675} \and  Diego Klabjan \inst{1} \orcidID{0000-0003-4213-9281}\and Jean Utke
\inst{2}\orcidID{0000-0002-3377-1990}}
\authorrunning{H. Li \etal}
%
\institute{Northwestern University, Evanston IL 60208, USA
\email{\{hanqingli2025@u.,d-klabjan@\}northwestern.edu}\\
\and
Allstate Insurance Company, Northbrook IL, 60062, USA\\
\email{jutke@allstate.com}}
\maketitle              
\begin{abstract}
This paper introduces a new, unsupervised method for automatic video summarization using ideas from generative adversarial networks but eliminating the discriminator, having a simple loss function, and separating training of different parts of the model. An iterative training strategy is also applied by alternately training the reconstructor and the frame selector for multiple iterations. Furthermore, a trainable mask vector is added to the model in summary generation during training and evaluation. The method also includes an unsupervised model selection algorithm. Results from experiments on two public datasets (SumMe and TVSum) and four datasets we created (Soccer, LoL, MLB, and ShortMLB) demonstrate the effectiveness of each component on the model performance, particularly the iterative training strategy. Evaluations and comparisons with the state-of-the-art methods highlight the advantages of the proposed method in performance, stability, and training efficiency.

\end{abstract}
\section{Introduction}
\label{sec:intro}
On the internet, there is a seemingly endless stream of social media and sharing platforms carrying a sea of video content, which creates the need to navigate and locate valuable clips efficiently. One solution to this need lies in video summarization. Video summarization aids in browsing large and continually growing collections by synthesizing an overwhelming amount of information into an easily digestible form. In this paper, we propose an unsupervised learning model that automatically summarizes video. We name the model SUM-SR according to its summarization function and architecture containing a selector and a reconstructor.

Most research approaches the video summarization task in a supervised manner, using ground-truth annotations to guide the learning process, Apostolidis \etal\cite{apostolidis2021video}. Nonetheless, there are also several unsupervised approaches that are trained without the need of ground-truth data, eliminating the need for laborious and time-consuming annotation tasks. The competitive performance of some unsupervised methods and the limited availability of ground-truth data suggest that unsupervised video summarization approaches have significant potential.

SUM-SR builds on SUM-GAN-AAE while removing the discriminator. SUM-SR consists of a selector for choosing key fragments from the original video and a reconstructor with an attention mechanism for reconstructing the original video from the video summary. However, instead of using an additional discriminator to compare the original video with a summary-based reconstructed version like other works\cite{apostolidis2019stepwise,apostolidis2020ac,apostolidis2020unsupervised,apostolidis2022summarizing,mahasseni2017unsupervised, gonuguntla2019enhanced,jung2019discriminative,jung2020global,kanafani2021unsupervised}, which increases the complexity of training, we directly calculate the mean square error (MSE) between the embeddings of the two videos as the loss function. Compared to numerous loss functions in SUM-GAN-AAE, SUM-SR  uses only the reconstruction and a regularization loss to guide the training process. We introduce an extra training step that separates the training of the reconstructor and the selector as they have different functionalities. Moreover, we extend this term-by-term training strategy to an iterative one that trains the two parts of the model alternately with multiple iterations. Such an approach further improves the model's performance. Lastly, we design an unsupervised algorithm to select the best model after training. We test the performance of the model on two benchmark datasets: TVSum, Song \etal\cite{tvsum}, and SumMe, Gygli \etal\cite{summe}, as well as four datasets we created. The proposed model demonstrates better performance than previous state-of-the-art methods by $8.5\%$ on average based on the per dataset best benchmark and $9.2\%$ based on a single best benchmark. The implementation and datasets are available at \url{https://github.com/hanklee97121/SUM-SR-5iter/tree/main}.

Our contributions are as follows.
\begin{itemize}
    \setlength\itemsep{0em}
  \item We create a new framework for the task of unsupervised video summarization by comparing a summary-based reconstructed video with its original video only through a reconstructor network without using a discriminator.
  \item We introduce an extra training step for the reconstructor and an iterative training strategy to increase the performance of the model.
  \item We also design a function to select the best model after training in an unsupervised manner.
\end{itemize}
The rest of the paper is organized as follows. In \cref{sec:related}, we review previous research on unsupervised video summarization. In \cref{sec:model}, we detail the proposed unsupervised deep learning approach. In \cref{sec:exp}, we present the experimental results and compare them to state-of-the-art methods. Finally, in \cref{sec:con}, we conclude the paper.

\section{Related Work}
\label{sec:related}
In recent years, there have been several approaches to automatic video summarization and related fields such as video highlight detection and video sementic compression. Highlight detection aims to extract brief video segments from unedited recordings that capture the user's primary focus or interest. Supervised approaches\cite{gygli2016video2gif, sun2014ranking, xu2021cross} predict fine-grained highlight scores, while unsupervised methods\cite{badamdorj2022contrastive, khosla2013large} identify highlight segments without human annotation. Recently, some studies\cite{Lin_2023_ICCV, Zala_2023_CVPR, Moon_2023_CVPR} have also employed text queries to locate desired highlight clips. In contrast to highlight detection, our work focuses on video summarization that generates a comprehensive summary of a video. Another closely related research area is video semantic compression. It focuses on reducing the size of digital video data while preserving essential semantic information necessary for downstream video analysis tasks\cite{tian2023non}. Tian \etal\cite{tian2023non} are the first to introduce this concept, proposing an unsupervised framework for video semantic compression and a special framework tailored for low-bitrate videos\cite{tian2024coding}. In contrast to video semantic compression, our research concentrates on generating concise video summaries for human comprehension rather than video analysis tasks. There are supervised and unsupervised video summarization methods based on the presence of ground-truth labels. In this section, we focus on presenting relevant papers on unsupervised approaches. If readers are interested in supervised video summarization, they are referred to the review by Apostolidis \etal\cite{apostolidis2021video}.

In unsupervised video summarization, the absence of ground-truth labels is addressed by focusing on key characteristics of effective summaries. Recent methods aim to create summaries that accurately represent the original content. These approaches typically employ Generator-Discriminator architectures and adversarial training to ensure the summarization component produces a summary that can reconstruct the original video effectively\cite{jung2020global,kanafani2021unsupervised,apostolidis2019stepwise}. Mahasseni \etal\cite{mahasseni2017unsupervised} introduced adversarial learning in video summarization by combining a Variational Auto-Encoder, a discriminator, and an LSTM-based keyframe selector. In SUM-GAN-AAE\cite{apostolidis2020unsupervised}, Apostolidis \etal replaced the Variational Auto-Encoder with a deterministic attention-based Auto-Encoder, while in AC-SUM-GAN\cite{apostolidis2020ac}, they embedded an Actor-Critic model to merge adversarial and reinforcement learning. In their latest work\cite{apostolidis2022summarizing}, Apostolidis \etal substituted the GAN with an attention mechanism that leverages frame uniqueness and diversity. Jung \etal\cite{jung2019discriminative}, building on Mahasseni \etal's model\cite{mahasseni2017unsupervised}, developed the Chunk and Stride Network (CSNet), which used both local and global temporal information and introduced a variance loss to highlight dynamic scenes.

Some other unsupervised video summarization methods use hand-crafted reward functions to quantify characteristics like representativeness and diversity, employing reinforcement learning for training\cite{gonuguntla2019enhanced,gonuguntla2019enhanced}. Zhou \etal\cite{zhou2018deep} used an LSTM-based architecture with rewards for diversity and representativeness, treating summarization as a sequential decision-making process. Zhao \etal\cite{zhao2019property} combined summarization and reconstruction, using reconstruction to assess how well the summary infers the original video. Yaliniz \etal\cite{yaliniz2021using} applied independent recurrent neural networks\cite{li2018independently} with rewards for representativeness, diversity, and temporal coherence. 

Compared to GAN-based methods mentioned above\cite{mahasseni2017unsupervised,apostolidis2019stepwise,apostolidis2020unsupervised,apostolidis2020ac,jung2019discriminative,jung2020global,kanafani2021unsupervised}, the proposed approach eliminates the discriminator, thereby simplifying the training steps and removing the potential risk of unstable training when using GANs, Zhou \etal\cite{zhou2018deep}. We do not update every trainable weight in the model at each epoch, but we separate the training of the selector and the reconstructor to enhance their performance. The reinforcement learning methods\cite{zhou2018deep,wang2016temporal,zhao2019property,yaliniz2021using,li2018independently,gonuguntla2019enhanced} above employ hand-crafted reward functions which are hard to tailor and often lead to poor performance, Apostolidis \etal\cite{apostolidis2020ac}. We let the model learn how to construct a good summary from the original video through a deep learning approach instead of optimizing hand-crafted reward functions through a reinforcement learning approach. The most recent research by Apostolidis \etal\cite{apostolidis2022summarizing} contains neither GANs nor reinforcement learning. Nevertheless, they train their model only with the average distance between frame-level importance scores and a regulation hyperparameter, which is not directly related to creating a good summary and causes unstable model performance. In contrast, we include the reconstruction loss in the training process, which is built upon the assumption that a good summary could help recover the video. All previous works\cite{mahasseni2017unsupervised,apostolidis2019stepwise,apostolidis2020unsupervised,jung2019discriminative,jung2020global,kanafani2021unsupervised,zhou2018deep,wang2016temporal,zhao2019property,yaliniz2021using,li2018independently,gonuguntla2019enhanced,apostolidis2020ac} select the best model based on its performance on the validation set, which needs true labels. We introduce a new method associated with the proposed model to select the best model in unsupervised fashion using only the reconstruction and sparsity losses on the validation dataset.

\section{Model}
\label{sec:model}
This section explains the design and structure of the SUM-SR model and the training process. We describe in detail the model selection method and the function that generates a video summary from the output (importance scores) of the selector.

\subsection{Model Structure}
\label{sec:ModelStru}
Following the problem setting in SUM-GAN\cite{mahasseni2017unsupervised}, we subsample each video and use a pre-trained CNN to encode each frame. Each video is represented by a sequence of vectors $V=(\boldsymbol{x}_1, \boldsymbol{x}_2, \dots, \boldsymbol{x}_n)$, where $\boldsymbol{x}_i\in\mathbb{R}^d$ is the embedding of the frame $i$ of video $V$. We treat the summarization task as a binary classification problem. For each frame, the model decides whether or not to include it in the summary and we view the probability of inclusion as the importance score for this frame.

\begin{figure}
    \centering
    \includegraphics[width=0.7\textwidth]{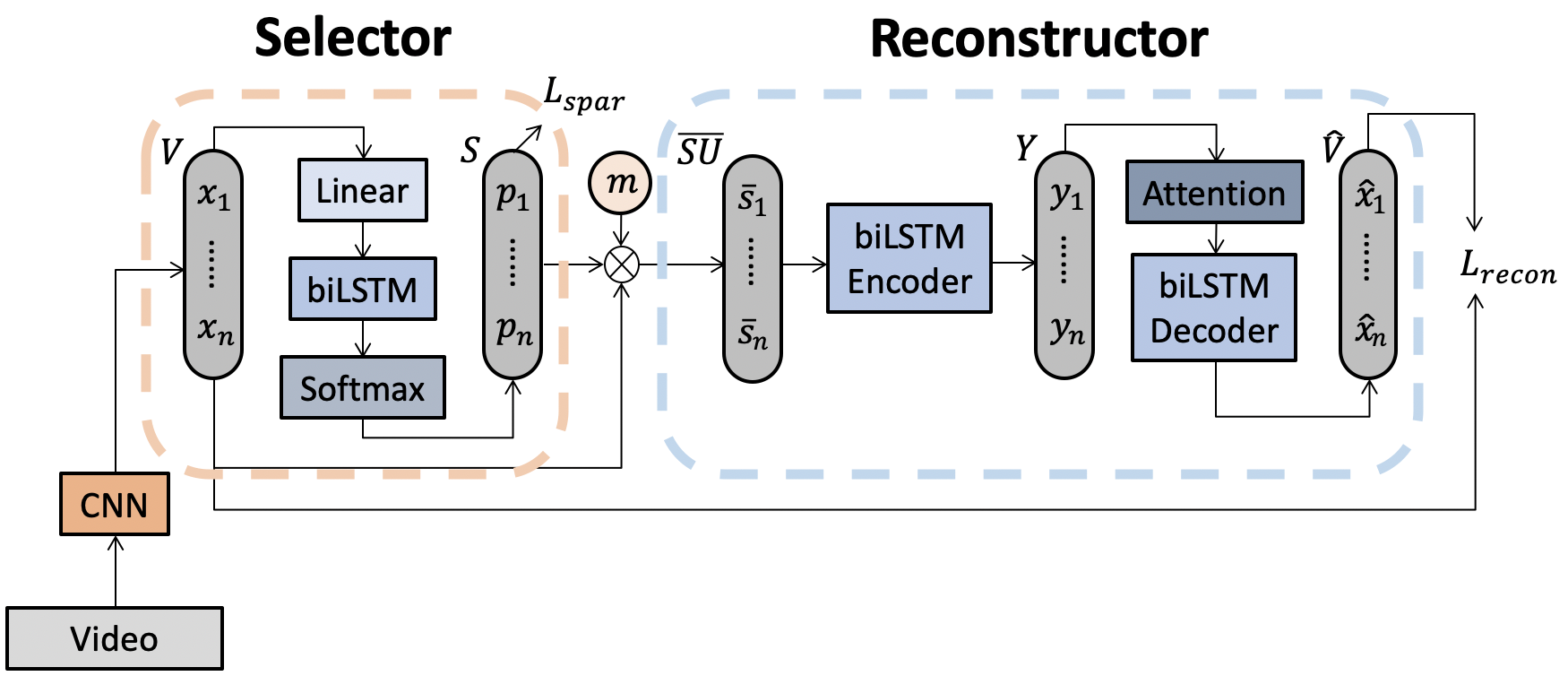}
    \caption{The proposed SUM-SR architecture.}
    \label{fig:archi}
\end{figure}

The proposed model is composed of a selector and a reconstructor (see \cref{fig:archi}), analogous to the selector and the reconstructor in SUM-GAN-AAE\cite{apostolidis2020unsupervised}. In the following sections, we denote the selector as $sNet$ and the reconstructor as $rNet$. The selector has a linear layer to compress the input dimension from $d$ to $d_h$, a bidirectional LSTM, and an output layer that maps the output $\boldsymbol{h}_{i}$ of the LSTM to an importance score $p_i$. The output layer first maps $\boldsymbol{h}_{i}$ to a two-dimensional vector $\boldsymbol{\hat{h}}_i$ and computes $p_i$ by softmax function with temperature $\tau$ as follows: 
\begin{align}
    \boldsymbol{h}_i &= \mbox{biLSTM}(\mbox{Lin}(\boldsymbol{x}_i), \boldsymbol{h}_{i-1})\\
    \boldsymbol{\hat{h}}_i &= \mbox{Lin}(\boldsymbol{h}_i)\\
    p_i &= \mbox{softmax}(\boldsymbol{\hat{h}}_i,\tau)_1.
\end{align}

Then we define $S = ( p_1, p_2, p_3, ..., p_n)$ as the importance scores of video $V$. Given video $V$ and importance scores $S$, we use a non-parametric function $f$ to create a summary by selecting important frames. The output is a vector $A = ( a_1, a_2, ..., a_n)$ with binary entries $a_i\in \{0, 1\}$ that indicate whether the $i$-th frame is selected or not. From $A$, we build a summary $SU = (\boldsymbol{s}_1, \boldsymbol{s}_2, \dots, \boldsymbol{s}_n)$, where $\boldsymbol{s}_i = \boldsymbol{x}_i$ if $a_i=1$ and $\boldsymbol{s}_i = \boldsymbol{m}$ if $a_i=0$. Vector $\boldsymbol{m}$ is a mask vector with dimension $d$. We explain more about $f$ and $\boldsymbol{m}$ in later sections.

Since the function $f$ is not differentiable, during training, we use a different method to create a trainable summary $\overline{SU} = (\boldsymbol{\bar{s}}_1, \boldsymbol{\bar{s}}_2, \dots, \boldsymbol{\bar{s}}_n)$ as in the paper by Mahasseni \etal\cite{mahasseni2017unsupervised}. Each entry $\boldsymbol{\bar{s}}_i$ is a weighted sum of $\boldsymbol{x}_i$ and $\boldsymbol{m}$ given by $\boldsymbol{\bar{s}}_i = p_i\cdot \boldsymbol{x}_i + (1-p_i)\cdot \boldsymbol{m}$.

The reconstructor of the model is an autoencoder with an attention block introduced in the work by Apostolidis \etal\cite{apostolidis2020unsupervised}. Both the encoder and decoder are bi-directional LSTMs with the input to the encoder being $\overline{SU}$. Focusing on the attention block (denoted as $m\_attn$), for any time step $i\in [2: n]$, the attention block has access to the encoder output $Y = (\boldsymbol{y}_1, \boldsymbol{y}_2, \dots, \boldsymbol{y}_n)$, where $\boldsymbol{y}_i\in \mathbb{R}^{d_h}$, and the previous hidden state of the decoder, $\boldsymbol{z}_{i-1}\in\mathbb{R}^{d_h}$. We compute the attention energy vector $\boldsymbol{e}_i\in\mathbb{R}^{n}$ from $Y$ and $\boldsymbol{z}_{i-1}$ by $\boldsymbol{e}_i = Y^{T}W_b\boldsymbol{z}_{i-1}$ where $W_b$ is a trainable matrix. At time step $i = 1$, we use the last hidden state of the encoder $\boldsymbol{h}_e$ to calculate $\boldsymbol{e}_1$. Afterward, we apply a softmax function on $\boldsymbol{e}_i$ to get a normalized attention weight vector $\boldsymbol{w}_i=\mbox{softmax}(\boldsymbol{e}_i)$ and multiply $\boldsymbol{w}_i\in\mathbb{R}^n$ with the encoder's output to produce a context vector $\boldsymbol{y'}_i = Y\boldsymbol{w}_i$. The context vector $\boldsymbol{y'}_i\in\mathbb{R}^{d_h}$ and the previous output of the decoder are concatenated together to form the input to the decoder at time step $i$. 
Given $\overline{SU}$ as the input, the reconstructor outputs the reconstructed video $\hat{V} = (\boldsymbol{\hat{x}}_1, \dots,\boldsymbol{\hat{x}}_n)$, $\boldsymbol{\hat{x}}_i\in\mathbb{R}^d$, as follows: 
\begin{align}
    Y &= Encoder(\overline{SU})\\
    \hat{V} &= Decoder_{atten}(Y).
\end{align} 
\noindent $Decoder_{atten}$ is a bi-directional LSTM with the attention block $m\_attn$.

During training, we use two loss functions introduced in SUM-GAN\cite{mahasseni2017unsupervised}: 1) reconstruction loss, $L_{recon}$, and 2) regularization loss, $L_{spar}$. Following SUM-GAN-AAE, Apostolidis \etal\cite{apostolidis2020unsupervised}, our goal is to train the selector to generate a summary that could be reconstructed to the original video through the reconstructor. We define the reconstruction loss as the Euclidean distance between the original video frame embeddings $V$ and the reconstructed video frame embeddings $\hat{V}$ based on $L_{recon} = ||V-\hat{V}||^2$. To avoid the trivial solution of selecting all frames, we introduce the Summary-Length regularization, Mahasseni \etal\cite{mahasseni2017unsupervised}, which penalizes the model when it assigns high importance scores to a large number of frames and introduces diversity in the video summary. The regularization loss is computed by $L_{spar} = ||\frac{1}{n}\sum_{i=1}^{n}p_{i}-\sigma||$, where $\sigma$ is a hyperparameter between 0 and 1. We train the model based on the loss function $L_{model} = L_{recon} + L_{spar}$.

\subsection{Training Strategy}
\label{sec:ModelTrain}
\begin{figure}
    \centering
    \includegraphics[width=0.7\textwidth]{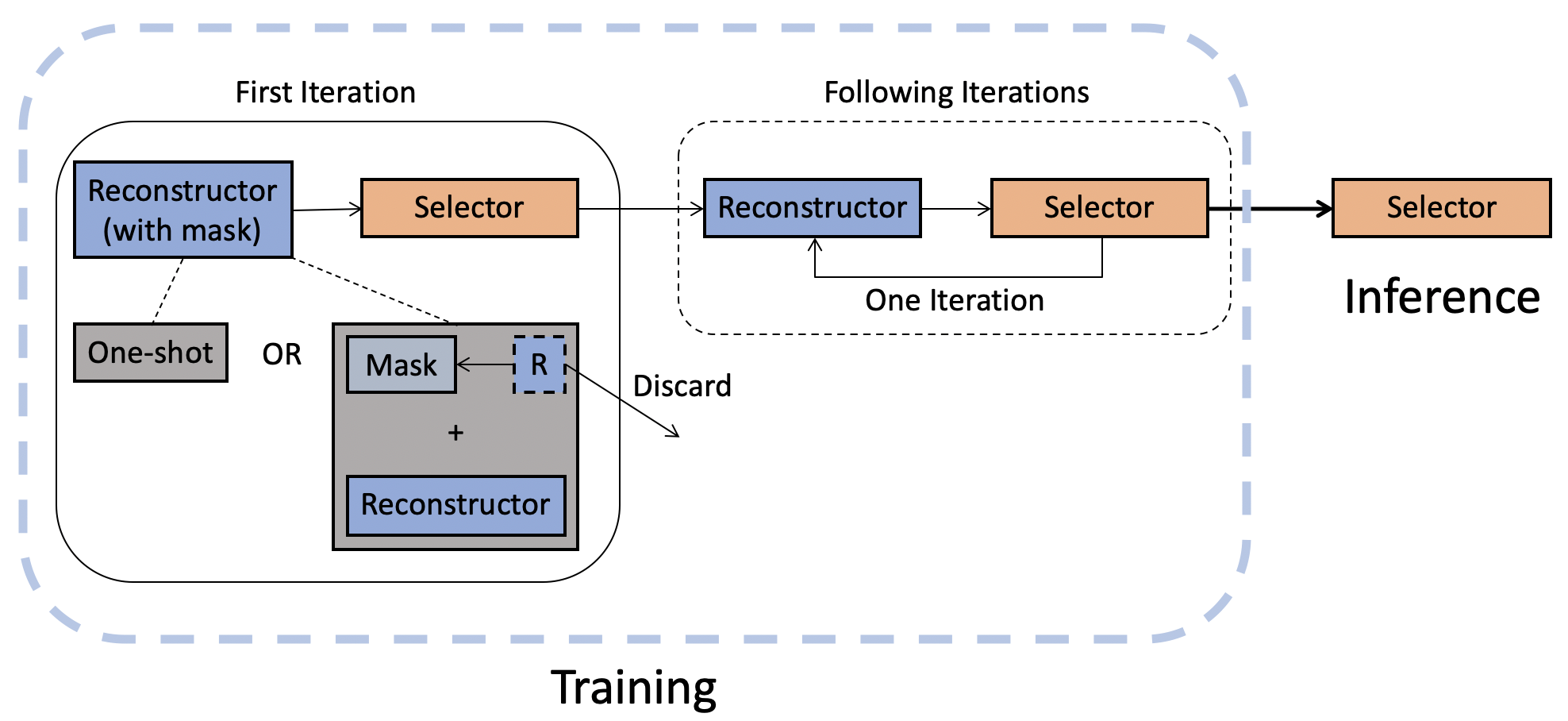}
    \caption{The training steps of SUM-SR. During training, an iteration includes one reconstruction and one selection. We train the mask vector only in the first iteration (if there are multiple iterations).}
    \label{fig:train}
\end{figure}

For inference, we keep only the selector to generate a video summary after training is complete. To make the training process more focused on updating the selector, we separate the training of the selector and the reconstructor. One iteration consists of training first only the reconstructor and then only the selector. We iterate several times. To prevent in the first iteration to train the selector with random reconstruction weights, we train the reconstructor first, see \cref{fig:train}.  Our goal is to create a shorter video summary with length $\alpha L$ from a video with length $L$, where $\alpha$ is the summary rate in $(0, 1)$, and the reconstructor aims to reconstruct the original video from the shorter video. Thus, when training the reconstructor, we create such a shorter video by randomly replacing some of the vectors $\boldsymbol{x}_i$ from a video $V=(\boldsymbol{x}_1, \boldsymbol{x}_2, \dots, \boldsymbol{x}_n)$ with a mask vector $\boldsymbol{m}$ to shorten the video length by masking information. Because our summary rate is $\alpha$, we want to keep $\alpha$ fraction of frames and replace the rest with the mask vector $\boldsymbol{m}$ to get $V'=(\boldsymbol{x'}_1, \boldsymbol{x'}_2, \dots, \boldsymbol{x'}_n)$, where $p(\boldsymbol{x'}_i=\boldsymbol{x}_i)=\alpha$, $p(\boldsymbol{x'}_i=\boldsymbol{m})=1-\alpha$. We feed $V'$ into the reconstructor to get a reconstructed video $\hat{V}' = (\boldsymbol{\hat{x}'}_1, \boldsymbol{\hat{x}'}_2, \dots, \boldsymbol{\hat{x}'}_n)$, and train the reconstructor by $L_{recon} = ||V-\hat{V}'||^2$. Then, we train only the selector based on $L_{model}$ in \cref{sec:ModelStru}.

The mask vector $\boldsymbol{m}$ is also trainable. To train the mask vector, we develop two strategies. The first strategy is updating the mask vector $\boldsymbol{m}$ together with the reconstructor when training only the reconstructor in the first iteration. Another strategy is to train the mask vector $\boldsymbol{m}$ alone first. We first create a new reconstructor $R$ (we use $R$ only to train the mask vector) and initialize $\boldsymbol{m}$ to zero vector. Then, we randomly replace some vectors in $V$ with $\boldsymbol{m}$ with probability $1-\alpha$ to get $V'$ and feed it into the reconstructor $R$. We train both  $R$ and $\boldsymbol{m}$ with the loss function $L_{mask} = \frac{1}{|{\cal D}|}\sum_{j\in {\cal D}}||\boldsymbol{x'}_j-\boldsymbol{\hat{x}'}_j||^2$, where $\cal D$ is the set of indices $j$ such that $\boldsymbol{x'}_j = \boldsymbol{m}$ and $\boldsymbol{\hat{x}'}_j$ is an output from $R$. After this training process, the mask vector remains fixed during the reconstructor's training step and the selector's training step (the rest of $R$ is discarded).

We define one reconstruction and selection as an iteration. To further improve the model performance, we train the model with multiple iterations. In each iteration, we initialize the model as the selected model from the previous iteration.

\subsection{Summarization}
\label{sec:ModelSum}
After obtaining importance scores $S$, we use a non-parametric function $f(V, S)$ to generate a summary of video $V$. We first obtain video shots (a continuous clip of a video that contains multiple frames) with the KTS algorithm introduced in the paper by Potapov \etal\cite{potapov2014category}. Then, we calculate the shot-level importance scores by averaging the frame-level importance scores of each shot. Finally, we generate the summary by maximizing the sum of the shot-level importance scores. Meanwhile, we ensure the summary length is shorter than $\alpha$ fraction of the original video length. We formulate this as the knapsack problem in the work by Gygli \etal\cite{summe}
\begin{equation}
    \max\limits_{\hat{A}}\sum_{i=1}^{N}\hat{a}_i\cdot \hat{p}_i,\ \text{s.t.} \sum_{i=1}^{N}\hat{a}_i\cdot l_i \leq \alpha\cdot L, \hat{a}_i \in \{0, 1\},
\end{equation}
\noindent
where $N$ is the number of shots, $L$ is the length of the original video $V$, and $\hat{p}_i$ is the shot-level importance score of the $i$-th shot. Binary variable $\hat{a}_i$ indicates whether to include the $i$-th shot in the summary, and $l_i$ is the length of the $i$-th shot. We define the shot-level summary vector as  $\hat{A}=( \hat{a}_1, \hat{a}_2, ..., \hat{a}_N )$. 

\subsection{Model Selection}
\label{sec:ModelSel}
Since the model is unsupervised, we need an unsupervised method to select the model for inference. In a single iteration, for all models from all training epochs, we generate the summary from the selector ($sNet_i$) according to \cref{sec:ModelStru} and obtain the reconstructed video using the reconstructor ($rNet_i$). Here $i=1, ..., E$ with $E$ being the total number of epochs. Consider epoch $i$ and video $V_j$ in the validation dataset. We first calculate $S_{ij}=sNet_i(V_j)$, $A_{ij}=f(V_j, S_{ij})$. By using $A_{ij}$, we next generate $SU_{ij}$ according to \cref{sec:ModelStru}. Note that this is different from training, where $\overline{SU}$ is used because of to the need for differentiability. Here, $SU_{ij}$ is the actual summary we construct from the model. Finally, we get the reconstructed video embedding by $\hat{V}_{ij} = rNet_0(SU_{ij})$. We use $rNet_0$ from the reconstructor-only training stage as it is expected to be of high quality. At this point, we have $L_{recon_{ij}} = ||\hat{V}_{ij}-V_j||^2$ and $L_{spar_{ij}}$.

We follow by first averaging all samples in validation to get $\bar{L}_{recon_{i}}$ and $\bar{L}_{spar_{i}}$, which are then separately scaled so that across all $i$ the losses are in $[0, 1]$ to get $\bar{L}_{recon_{i}}^{normal}$ and $\bar{L}_{spar_{i}}^{normal}$. For epoch $i$, the selected model is 
\begin{equation}
    arg\underset{i}{max}(\bar{L}_{recon_{i}}^{normal}-\bar{L}_{spar_{i}}^{normal}).
    \label{equ:argmax}
\end{equation}
\noindent A more detailed rationale and discussion regarding this selection methodology can be found in Appendix A.

For the experiment without separating training of the reconstructor and selector, we first pick a reconstructor with the smallest validation reconstruction loss following the same expressions except that $rNet_i$ replaces $rNet_0$. This yields the best reconstruction model $\beta$. Finally, we repeat the previous model selection steps by using $rNet_{\beta}$ instead of $rNet_0$.

For the experiment with multiple iterations, we first pick a target model in each iteration using the aforementioned model selection method and pick the target model with the smallest reconstruction loss on the validation set.

\section{Experiments}
\label{sec:exp}
We select CA-SUM\cite{apostolidis2022summarizing}, AC-SUM-GAN\cite{apostolidis2020ac}, CSNet\cite{jung2019discriminative} and SUM-GAN-AAE\cite{apostolidis2020unsupervised} as benchmarks for performance comparison. According to previous works\cite{apostolidis2022summarizing,jung2019discriminative,apostolidis2020ac}, CA-SUM performs the best on TVSum, CSNet performs the best on SumMe and AC-SUM-GAN is the best approach using reinforcement learning. We also include SUM-GAN-AAE\cite{apostolidis2020unsupervised} because our approach builds on it. All benchmarks use ground truth summary in model selection, but our approach uses the unsupervised model selection method. For a fair comparison, we use an unsupervised model selection method for each benchmark model. For CA-SUM\cite{apostolidis2022summarizing}, we use its proposed unsupervised model selection method by choosing the model with the smallest model loss $L_{reg}$, as defined in \cite{apostolidis2022summarizing}, on the validation set. For AC-SUM-GAN\cite{apostolidis2020ac}, Apostolidis \etal mention a model selection method selecting the best model with the highest reward and simultaneously the smallest actor’s loss on the validation set. We follow this model selection method in our subsequent experiments with AC-SUM-GAN. CSNet\cite{jung2019discriminative} and SUM-GAN-AAE\cite{apostolidis2020unsupervised} do not have an unsupervised model selection method. Since their training strategies are similar to that of CA-SUM, following the model selection method of CA-SUM, we select the best model with an unsupervised method for SUM-GAN-AAE and CSNet by choosing the best model with minimum model loss on the validation set ($L_{model}=L_{sparsity}+L_{recon}$ for SUM-GAN-AAE and $L_{model}=L_{recon} + L_{prior} + L_{sparsity} + L_{v}$ for CSNet). Meanwhile, to better assess a model's performance, we run each model five times on each dataset with different random seeds and report the average.

\subsection{Datasets}
\label{sec:expData}
We evaluate the performance of the model on two public datasets and four datasets we created. The two public datasets are SumMe, Gygli \etal\cite{summe}, and TVSum, Song \etal\cite{tvsum}. The four datasets we created are Soccer, LoL, MLB, and ShortMLB, where each video is labeled by only one summary.
\begin{itemize}
    \item \textbf{SumMe}: It contains 25 videos with diverse contents (\eg, scuba diving, cooking, cockpit landing) from 1 minute to 6 minutes, captured from both moving and static views. Each video has been annotated by 15 to 18 keyframe fragments generated by different human evaluators. The average summary length is from $10.7\%$ to $15.5\%$ of the original video length. There are 20 training videos and 5 testing videos as in previous approaches\cite{apostolidis2020unsupervised,apostolidis2019stepwise,apostolidis2020ac,apostolidis2021video,apostolidis2022summarizing}.

    \item \textbf{TVSum}: It includes 50 videos of various types, such as news, vlogs, and documentaries, with lengths ranging from 1 to 11 minutes. Each video has been evaluated by 20 different human evaluators, who assigned a score of 1 to 5, with 1 indicating not important and 5 indicating very important, to every 2-second shot in the video. We use 40 videos for training and 10 videos for testing according to previous works\cite{apostolidis2020unsupervised,apostolidis2019stepwise,apostolidis2020ac,apostolidis2021video,apostolidis2022summarizing}.

    \item \textbf{Soccer}: It consists of 69 videos clipped from 11 soccer games where the video length ranges from 2 to 11 minutes. Nine videos are in the test set, and the other 60 videos are split into 50 training videos and 10 validation videos. Each video in the test set has a goal, which we label as a ground-truth summary.
    
    \item \textbf{LoL}: It comprises 55 videos extracted from 19 League of Legends matches, with lengths varying from 2 to 10 minutes. Out of these, 5 videos are designated for testing, while the remaining 50 videos are divided into 40 for training and 10 for validation. The ground-truth summary for each video in the test set is composed of segments related to the killing of a hero or the destruction of a tower.
    
    \item \textbf{MLB}: It has 60 videos from 5 MLB games, with durations between 5 and 10 minutes. Ten of these videos are selected for testing, and the remaining 50 are divided into 40 for training and 10 for validation. The ground-truth summary for each video in the test set is determined by frames that corresponding to a hit.
    
    \item \textbf{ShortMLB}: This dataset is a shorter version of MLB. We create ShortMLB by clipping each video in MLB to only 2 to 4 minutes. Thus, except for video length, the rest of this dataset is the same as MLB.
\end{itemize}
 
We create five random train-test splits for TVSum and SumMe following previous approaches\cite{apostolidis2019stepwise,mahasseni2017unsupervised,apostolidis2020unsupervised,apostolidis2020ac}, and five random train-test splits for Soccer, LoL, MLB, and ShortMLB.

\subsection{Evaluation}
\label{sec:expEva}
Following the previous approach by Zhang \etal\cite{zhang2016video}, we calculate the F-score to evaluate the quality of the summary generated by the model. 

For a single video, we compare the model-generated summary with user-generated summaries (for TVSum and SumMe) or the ground-truth summary (for Soccer, LoL, MLB, and ShortMLB) by computing the F-score for each pair of compared summaries. This F-score is the final F-score for this video for Soccer, LoL, MLB, and ShortMLB datasets. Each video in TVSum and SumMe has multiple user-generated summaries and thus has multiple F-scores. According to the study of SumMe and TVSum by Apostolidis \etal\cite{apostolidis2019stepwise}, there is no ideal summary that exhibits significant overlap with all annotators’ preferences in SumMe. Moreover, based on the consistency analysis for SumMe and TVSum by Gygli \etal\cite{summe} and Song \etal\cite{tvsum}, user-generated summaries in TVSum are more consistent for a single video than those in SumMe. Therefore, following the evaluation criteria in the work by Zhang \etal\cite{zhang2016video}, we take the maximum of the multiple F-scores to access the model performance for SumMe and the average of the multiple F-scores for TVSum. We report the average performance over all splits for each dataset.

\subsection{Implementation Details}
\label{sec:expImp}
Following the paper by Mahasseni \etal\cite{mahasseni2017unsupervised}, we subsample each video to 2fps and embed each frame to a vector of size $d=1024$ using GoogLeNet introduced by Szegedy \etal\cite{szegedy2015going} and trained on the ImageNet dataset. We set the regularization factor $\sigma=0.7$, the temperature $\tau = 0.5$, and the summary rate $\alpha=0.15$. All bidirectional LSTMs in the model have two layers with the hidden dimension $d_h=512$. The linear layer in the selector has the input dimension $d=1024$ and output dimension $d_h=512$. During training with Adam, we set the learning rate to $0.0001$ and the gradient clipping range to $[-5, 5]$. We initialize the model weights randomly. In one iteration, we first train the reconstructor for 100 epochs and then train the selector for 100 epochs.

We propose five versions of the proposed model. Each version has a unique training strategy as follows.
\begin{itemize}
    \item \textbf{SUM-SR}: We train the proposed model for 100 epochs without separating the reconstructor and the selector. The mask vector $\boldsymbol{m}$ is a constant zero vector. This corresponds to one iteration in \cref{fig:train} with training the reconstructor and selector together.
    \item \textbf{SUM-SR$_{\mathbf{sep}}$}: We train the reconstructor and the selector separately for 100 epochs but leave $\boldsymbol{m}$ as a zero vector. This corresponds to one iteration in following iterations in \cref{fig:train}.
    \item \textbf{SUM-SR$_{\mathbf{sepMa}}$}: We first train the mask vector $\boldsymbol{m}$ with the reconstructor for 100 epochs, followed by training the selector only for 100 epochs. This corresponds to the first iteration with one-shot training in \cref{fig:train}.
    \item \textbf{SUM-SR$_{\mathbf{sep-Ma}}$}: We separate the training of the mask vector $\boldsymbol{m}$ from the training of the reconstructor. We first train the mask vector for 100 epochs. Then, we train the reconstructor for 100 epochs. Finally, we train the selector for 100 epochs. This corresponds to the first iteration with "mask + reconstructor" training in \cref{fig:train}.
    \item \textbf{SUM-SR$_{\mathbf{5iter}}$}: We apply the iterative training strategy to SUM-SR$_{sepMa}$ for five iterations. We only update the mask vector $\boldsymbol{m}$ in the first iteration. This corresponds to the entire training part in \cref{fig:train}.
\end{itemize}

We train on NVIDIA GPU cards A100-PCIE-40GB GPU, GeForce GTX 1080, and GeForce RTX 2080 Ti. We used PyTorch version 1.0.1 with Python 3.6 as the development framework.

\subsection{Results}
\label{sec:expRes}
\begin{table}[tb]
  \caption{Comparison (F-score ($\%$)) of the proposed approach and state-of-the-art methods of unsupervised video summarization.}
  \label{tab:Comparsion}
  \centering
  \begin{tabular}{@{}lllllll@{}}
    \toprule
    Method & SumMe & TVSum & Soccer & LoL & MLB & ShortMLB\\
    \midrule
    SUM-GAN-AAE\cite{apostolidis2020unsupervised} & 46.81 &57.61 &21.06 &15.08 &15.13 &19.8 \\
    CSNet\cite{jung2019discriminative} & 44.61 &55.33 &20.94 &14.55 &16.3 & 19.11\\
    AC-SUM-GAN\cite{apostolidis2020ac} & 45.28 &57.98 &21.00 &14.95 & 17.09& 20.24\\
    CA-SUM\cite{apostolidis2022summarizing} & 45.07 &58.36 &21.70 &15.15 &17.68 & 20.18\\
    \midrule
    SUM-SR$_{\mathbf{5iter}}$ & \bf{51.26} &\bf{60.2} &\bf{23.84} &\bf{15.39} &\bf{19.38} & \bf{23.63}\\
  \bottomrule
  \end{tabular}
\end{table}

\begin{table}[tb]
  \caption{SUM-SR$_{5iter}$'s relative improvement (in percentage) on each dataset compared to the underlying method. We also calculate the average improvement based on the per dataset best bench-
mark and the single best benchmark CA-SUM.}
  \label{tab:relative}
  \centering
  \begin{tabular}{@{}lllllll|l}
    \toprule
    Method & SumMe & TVSum & Soccer & LoL & MLB & ShortMLB & Average\\
    \midrule
    SUM-GAN-AAE\cite{apostolidis2020unsupervised} & \bf{9.5\%} &4.5\% &13.2\% &2.1\% &28.1\% &19.3\%& 12.8\%\\
    CSNet\cite{jung2019discriminative} & 14.9\% &8.8\% &13.8\% &5.8\% &18.9\% & 23.65\%&14.3\%\\
    AC-SUM-GAN\cite{apostolidis2020ac} & 13.2\% &3.8\% &13.5\% &2.9\% &13.4\% & 17.2\%&10.7\%\\
    CA-SUM\cite{apostolidis2022summarizing} & 13.7\% &\bf{3.2\%} &\bf{9.9\%} &\bf{1.6\%} &\bf{9.6\%} & \bf{17.1\%}&\bf{9.2\%}\\
    \midrule
    Best & 9.5\% &3.2\% &9.9\% &1.6\% &9.6\% & 17.1\%&8.5\%\\ 
  \bottomrule
  \end{tabular}
\end{table}

\begin{figure}[tb]
    \centering
    \includegraphics[width=0.7\textwidth]{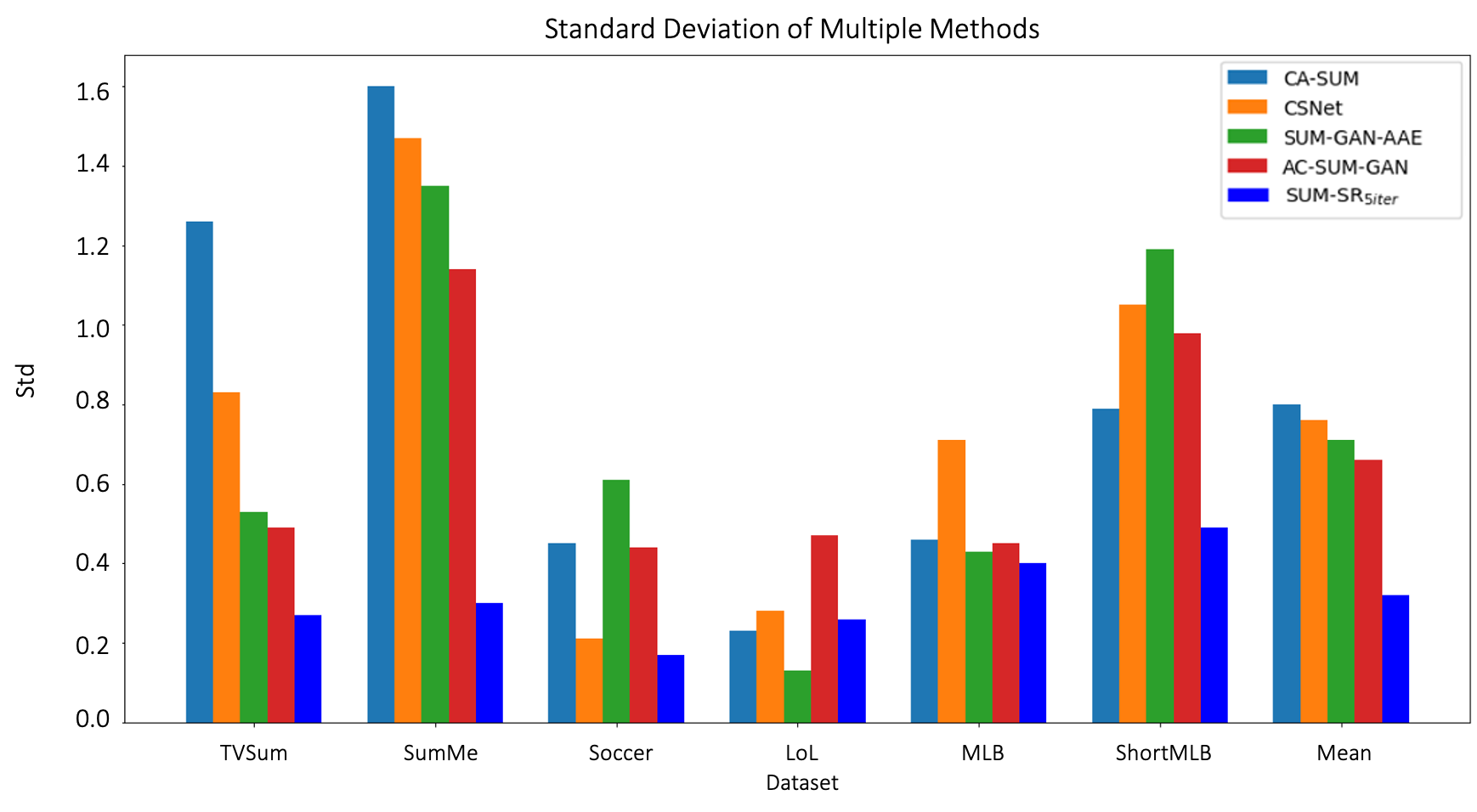}
    \caption{Comparison (standard deviation of the F-score) of different methods running multiple times with different random seeds on six datasets.}
    \label{fig:My_Model_std}
\end{figure}
We compare SUM-SR$_{\mathbf{5iter}}$, our best performer, with state-of-the-art unsupervised video summarization approaches. The results in \cref{tab:Comparsion} and \cref{tab:relative} indicate that SUM-SR$_{5iter}$ performs the best on all datasets, outperforming the best benchmark model CA-SUM by $9.2\%$ on average and per dataset best benchmark by $8.5\%$. The values on SumMe and TVSum of the benchmarks are aligned with those reported in the corresponding papers. The proposed training strategy effectively improves the model's summarization ability. Moreover, compared to other GAN-based methods, removing the discriminator has minimal effect on model performance.

Since we run each method with different random seeds several times, we investigate each model's stability by computing the final F-score's standard deviation over the different seeds.  According to \cref{fig:My_Model_std}, SUM-SR$_{5iter}$ has the smallest standard deviation compared to other methods on all except one dataset. Although SUM-GAN-AAE and CA-SUM have slightly smaller standard deviations on LoL, they have much higher standard deviations than the proposed model on other datasets. Comparatively, SUM-SR$_{5iter}$ is more resilient to randomness in training.

\begin{figure}[tb]
    \centering
    \begin{subfigure}[t]{0.3\textwidth}
        \includegraphics[width=\textwidth]{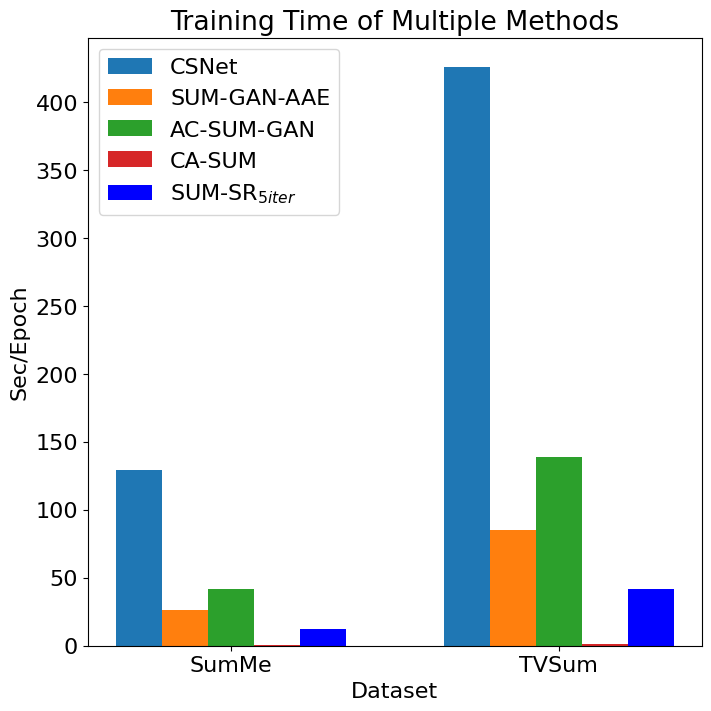}
        \caption{Per epoch training time.}
        \label{fig:PerRun}
    \end{subfigure}
    \begin{subfigure}[t]{0.3\textwidth}
        \includegraphics[width=\textwidth]{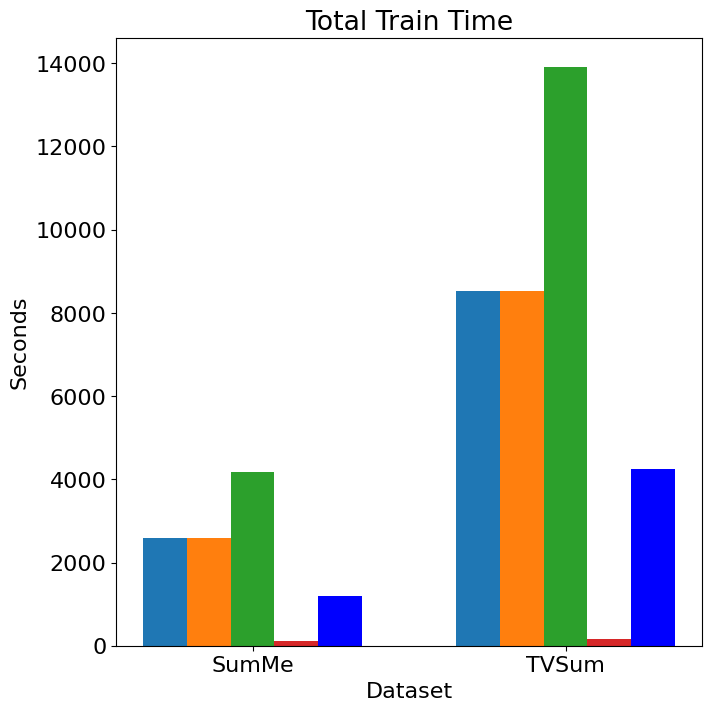}
        \caption{Total training time.}
        \label{fig:TotalRun}
    \end{subfigure}
    \begin{subfigure}[t]{0.3\textwidth}
        \includegraphics[width=\textwidth]{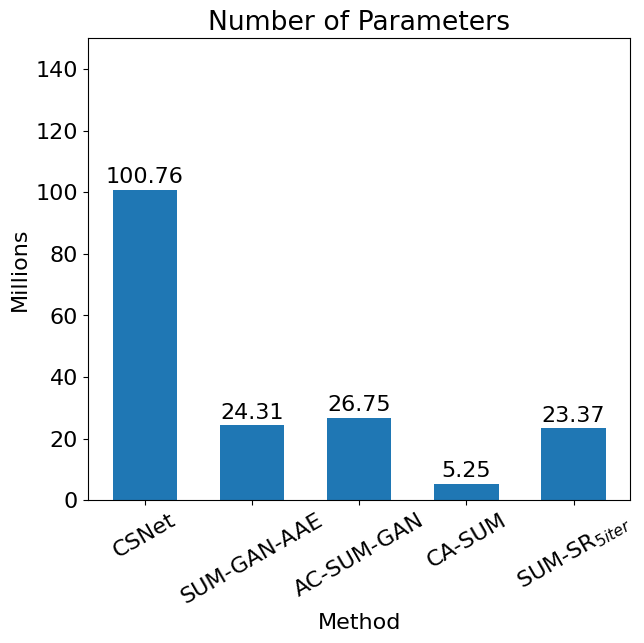}
        \caption{Model size.}
        \label{fig:size}
    \end{subfigure}

    \caption{Comparison of per epoch training time (sec/epoch), total training time (seconds) and number of parameters (millions) of different methods in the same computing environment.}
    \label{fig:time}
\end{figure}

Moreover, we compare the proposed approach with other unsupervised methods concerning the model size and the training time on SumMe and TVSum. Since different models train for different number of epochs (CSNet for 20 epochs, SUM-GAN-AAE and AC-SUM-GAN for 100 epochs and CA-SUM for 400 epochs), we calculate and compare the per epoch training time of different methods. We run each model in the same computing environment A100-PCIE-40GB over the same five data splits. The results in \cref{fig:time} show that SUM-SR$_{5iter}$ is smaller in size than other GAN-based models and trains much faster except CA-SUM. The total training time is 1,203 seconds on SumMe and 4,241 seconds on TVSum. Each training step takes  approximately half of the total training time. Eliminating the discriminator simplifies the training step and improves the training efficiency without a performance drop. On the other hand, even though CA-SUM is smaller and trains faster, its performance is worse by $9.2\%$ on average and more unstable than the proposed method. We also calculate the overall run time by multiplying the per epoch training time with the number of training epochs in \cref{fig:TotalRun}. Compared to other GAN-based methods, SUM-SR$_{5iter}$ requires notably less training time. Removing the discriminator decreases the total training time significantly.

For longer videos (20 minutes to an hour), we conducted experiments on a private dataset; therefore, the results are not included herein. Based on these experiments, we identify two effective approaches for applying our model to longer videos. The first approach is to decrease the frame sampling frequency to 1 fps or lower, as video content often remains consistent over several seconds. The second approach involves dividing the video into multiple shots using shot boundary detection methods, such as those applied in egocentric videos\cite{7750564}. Each shot is summarized individually, and these summaries are concatenated before applying the summarization model again to produce a comprehensive summary of the entire video.

\subsection{Ablation and Sensitivity Studies}
\label{sec:expAbla}
\begin{figure}[tb]
\centering
\includegraphics[width=0.8\linewidth]{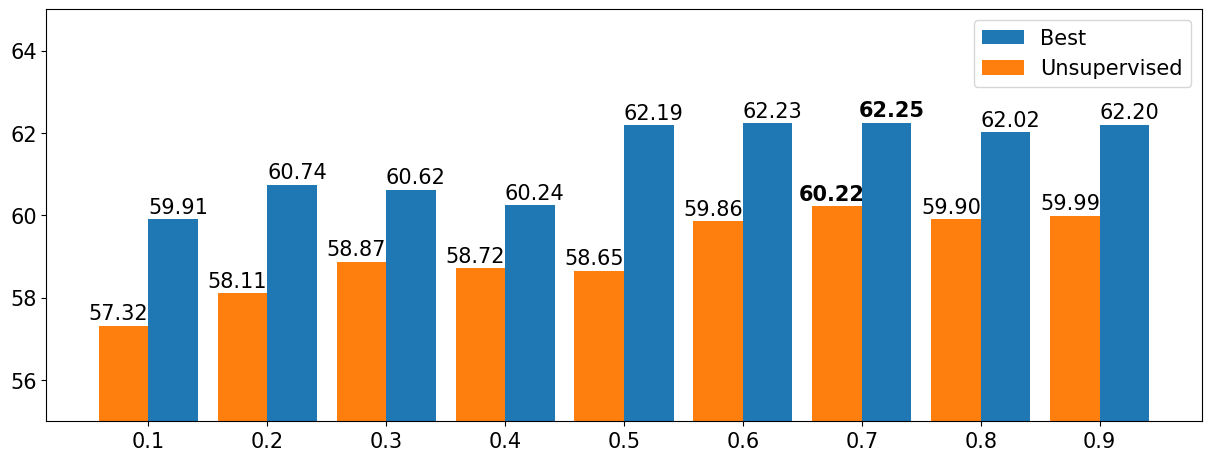}
\caption{Comparison (F-score ($\%$)) of different $\sigma$ in SUM-SR$_{sepMa}$ on the TVSum dataset with both unsupervised and supervised (best) model selection methods.}
\label{fig:sigma}
\end{figure}

We run sensitivity of the regularization hyperparameter $\sigma$ and the model versions. To explore the effect of $\sigma$, we run one version of the model (SUM-SR$_{sepMa}$) on TVSum with different $\sigma$ values from $0.1$ to $0.9$ and report both the model selected by our method and the best model. The best model is the model with best performance on test. According to \cref{fig:sigma}, the best option of $\sigma$ is 0.7. The increment of the $\sigma$ value does not always lead to performance improvement, but it is evident that models with high $\sigma$ values (0.6, 0.7, 0.8, 0.9) have better performance than those with low $\sigma$ values (0.1, 0.2, 0.3, 0.4). There is also a sudden increment in the F-score when $\sigma$ changes from 0.5 to 0.6 with the unsupervised selection method. The difference between the two extreme $\sigma$ is not remarkable attesting that the model is robust with respect to $\sigma$, yet it is beneficial to tune it.

\begin{table}[tb]
  \caption{Comparison (F-score ($\%$)) of SUM-SR$_{5iter}$ and other variations of the model with no iteration on the six datasets.}
  \label{tab:My_Model}
  \centering
  \begin{tabular}{@{}lllllll@{}}
    \toprule
    Method & SumMe & TVSum & Soccer & LoL & MLB & ShortMLB\\
    \midrule
    SUM-SR & 48.71 &59.08 &24.2 &15.31 &15.22 &19.9 \\
    SUM-SR$_{sep}$ & 49.39 &59.62 &24.63 &15.24 &15.39 & 20.31\\
    SUM-SR$_{sepMa}$ & 48.79 &59.83 &\bf{24.71} &15.3 &15.4 & 20.79\\
    SUM-SR$_{sep-Ma}$ & 48.51 &59.3 &24.37 &15.29 &15.09 & 20.52\\
    \midrule
    SUM-SR$_{\mathbf{5iter}}$ & \bf{51.26} &\bf{60.2} &23.84 &\bf{15.39} &\bf{19.38} & \bf{23.63} \\
  \bottomrule
  \end{tabular}
\end{table}

According to \cref{tab:My_Model}, SUM-SR$_{sepMa}$ outperforms other versions without iteration on most datasets except SumMe and LoL. On these two datasets, SUM-SR$_{sepMa}$ is the second-best model among variations without iteration. The performance of SUM-SR$_{sepMa}$ suggests that separating the training of the model and using a trainable mask vector $\boldsymbol{m}$ positively affect model performance. However, isolating the training of the mask vector $\boldsymbol{m}$ does not help the model to perform better. Thus, we test the iterative training strategy on SUM-SR$_{sepMa}$. The results show that the iterative strategy (SUM-SR$_{5iter}$) further improves the performance of SUM-SR$_{sepMa}$ on five out of six datasets but degrades the performance slightly on Soccer. Overall, SUM-SR$_{5iter}$ is the best version of the proposed method. We include further analysis of iterations in Appendix B and examples of video summaries in Appendix C. 

\section{Conclusion}
\label{sec:con}
We present a video summarization model that utilizes an autoencoder for unsupervised training and a part-by-part training strategy for performance improvement. Building on SUM-GAN-AAE,  Apostolidis \etal\cite{apostolidis2020unsupervised}, we create a variation that removes the discriminator and separates the training of the selector and the reconstructor. We also explore the iterative training method that trains the model with multiple iterations. Experiments on two public datasets (SumMe and TVSum) and four datasets of ourselves (Soccer, LoL, MLB, ShortMLB) show that removing the discriminator does not impair the model performance but decreases the model size and the training time. The proposed training strategy notably improves the model performance and makes the model outperform the best state-of-the-art method by $9.2\%$ and on per dataset best benchmark by $8.5\%$ on average. 
%
%
%
\bibliographystyle{splncs04}
\bibliography{main}

\newpage
\appendix
\section{Model Selection Method}
\label{app:model_select}
\begin{figure}
    \centering
    \begin{subfigure}[t]{0.25\textwidth}
        \includegraphics[width=\textwidth]{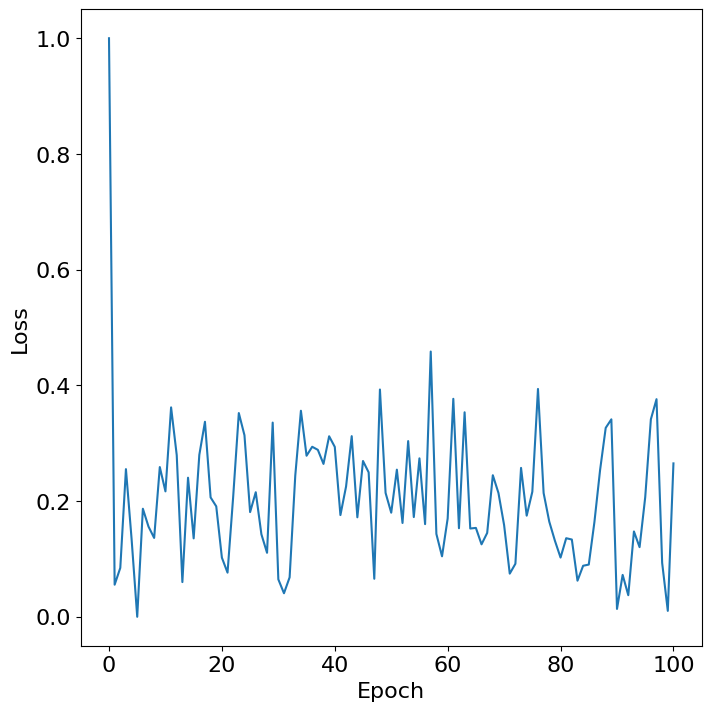}
        \caption{$\bar{L}_{recon_{i}}^{normal}$}
        \label{ReconNorm}
    \end{subfigure}
    \begin{subfigure}[t]{0.25\textwidth}
        \includegraphics[width=\textwidth]{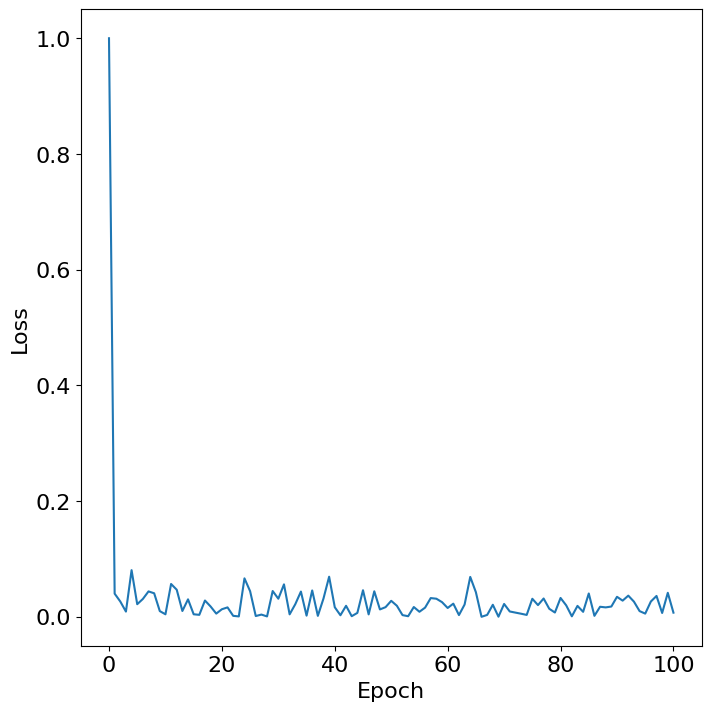}
        \caption{$\bar{L}_{spar_{i}}^{normal}$}
        \label{SparNorm}
    \end{subfigure}
    \begin{subfigure}[t]{0.25\textwidth}
        \includegraphics[width=\textwidth]{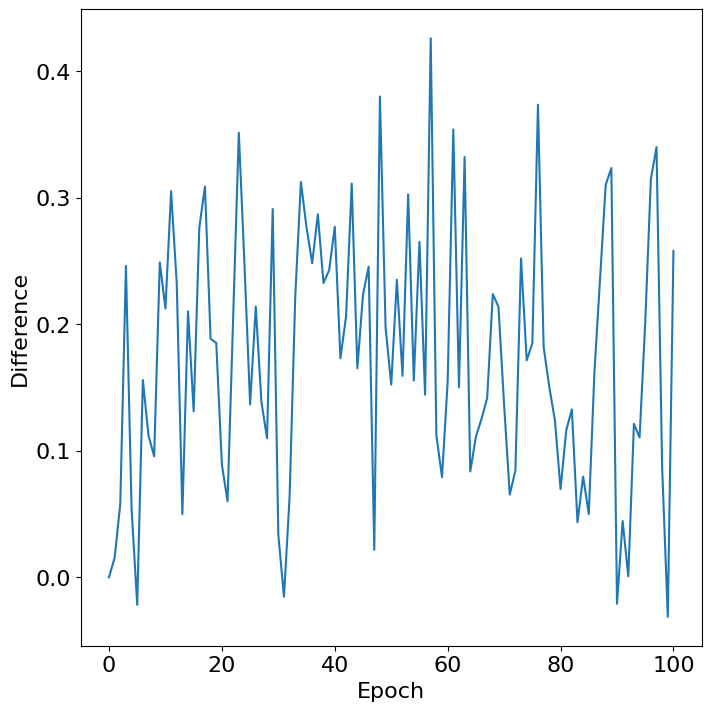}
        \caption{$\bar{L}_{recon_{i}}^{normal}-\bar{L}_{spar_{i}}^{normal}$}
        \label{DiffNorm}
    \end{subfigure}
    \caption{Normalized Loss Curve of one Experiment on TVSum}
    \label{LossC}
\end{figure}

The model selection method is decided based on experiment results and inspired by loss curves in \cref{LossC}. After normalizing both $\bar{L}_{recon_{i}}$ and $\bar{L}_{spar_{i}}$ to [0, 1], it is noteworthy that their difference is small because their initial values are large. Subsequently, as both losses converge towards a small value and oscillate in proximity to this converged state, the normalized difference starts increasing with oscillation. Based on the selection method, a model is selected after convergence, avoiding the trivial solution of a model with an extremely high reconstruction loss coupled with a minimal sparsity loss. The experiment result of the model performance in Section 4 of the main paper also proves the solidity of the model selection method.

\section{Impact of Iterations}
\label{app:iter}
\begin{figure}[tb]
    \centering
    \begin{subfigure}[t]{0.3\textwidth}
        \includegraphics[width=\textwidth]{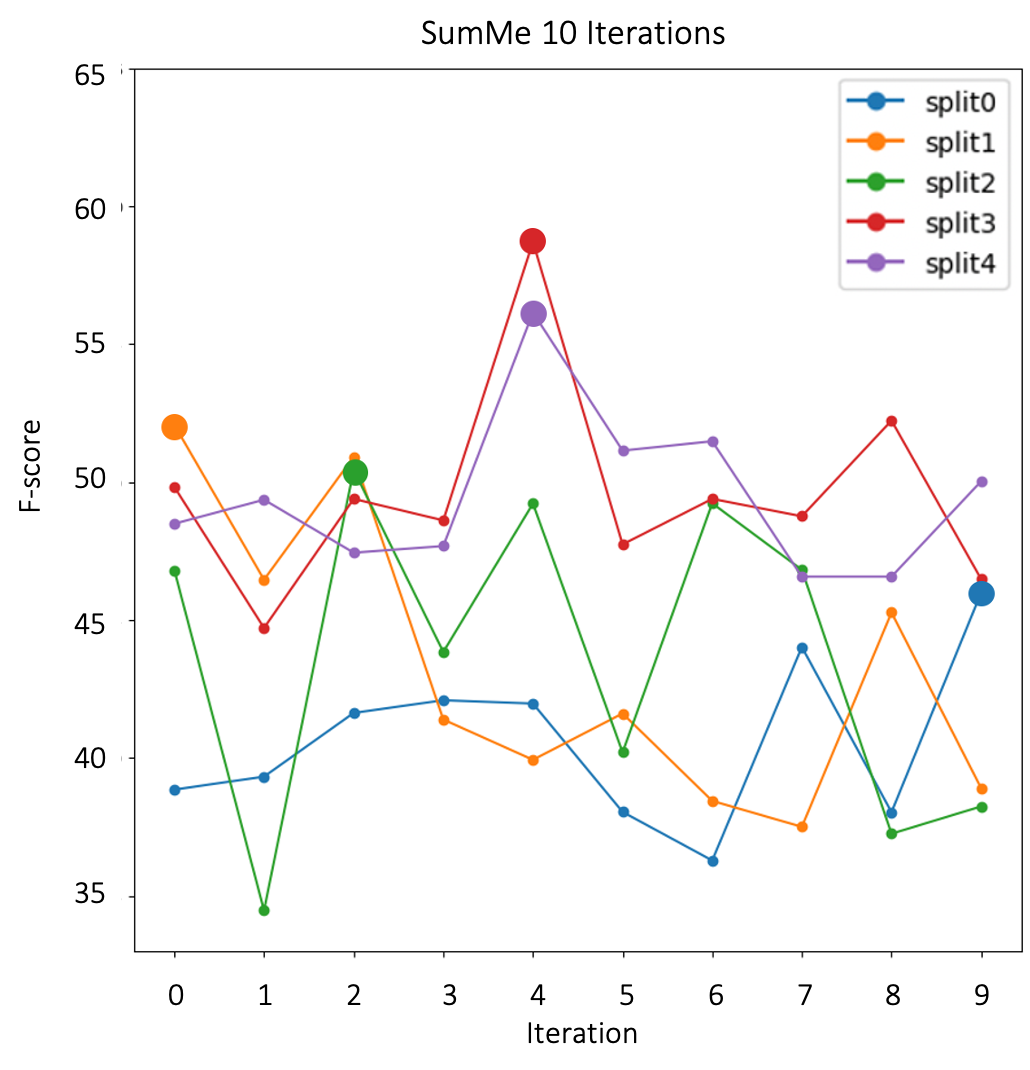}
        \caption{SumMe.}
    \end{subfigure}
    \begin{subfigure}[t]{0.35\textwidth}
        \includegraphics[width=\textwidth]{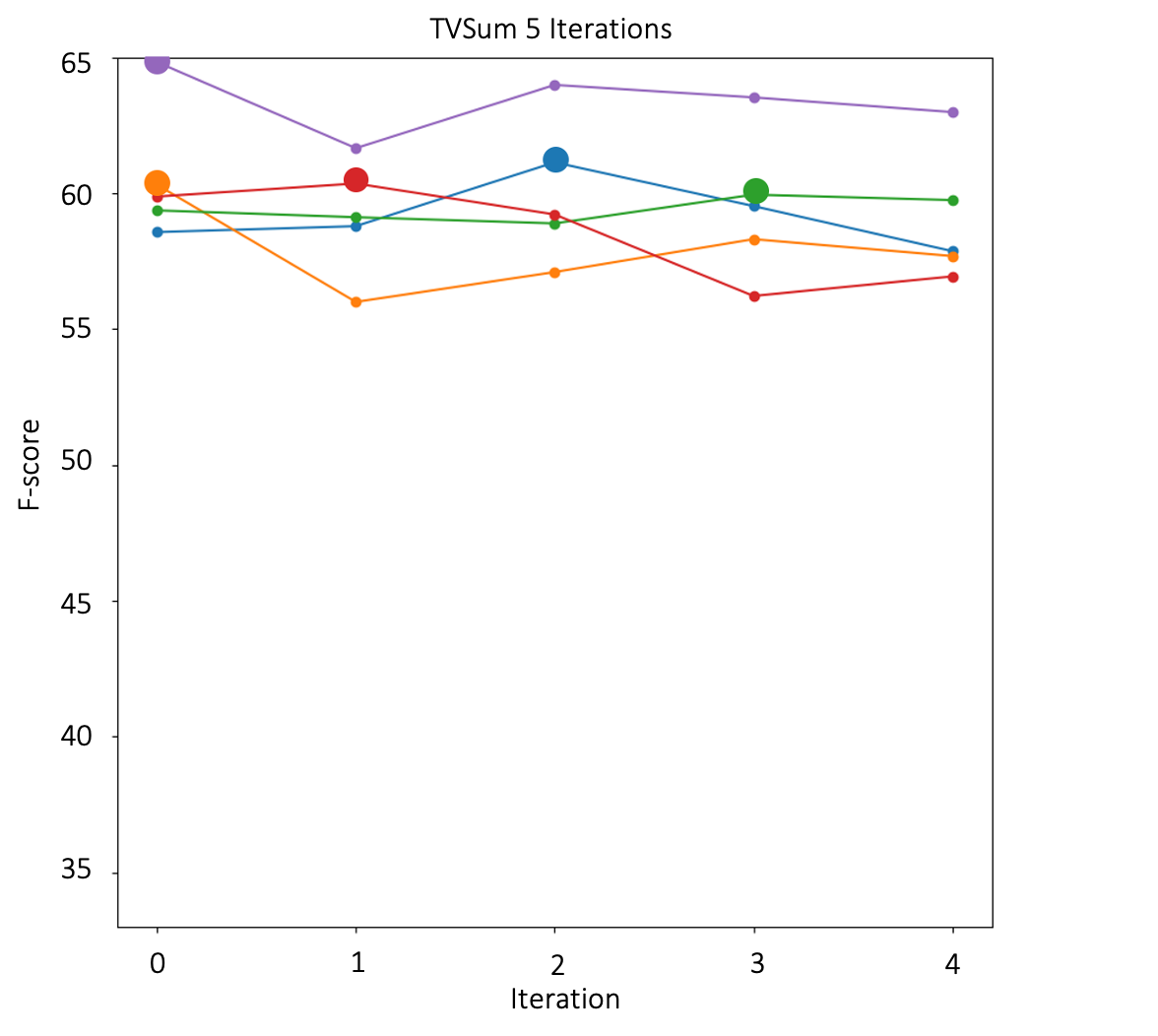}
        \caption{TVSum.}
    \end{subfigure}
    
    \bigskip
    \begin{subfigure}[b]{0.23\textwidth}
        \includegraphics[width=\textwidth]{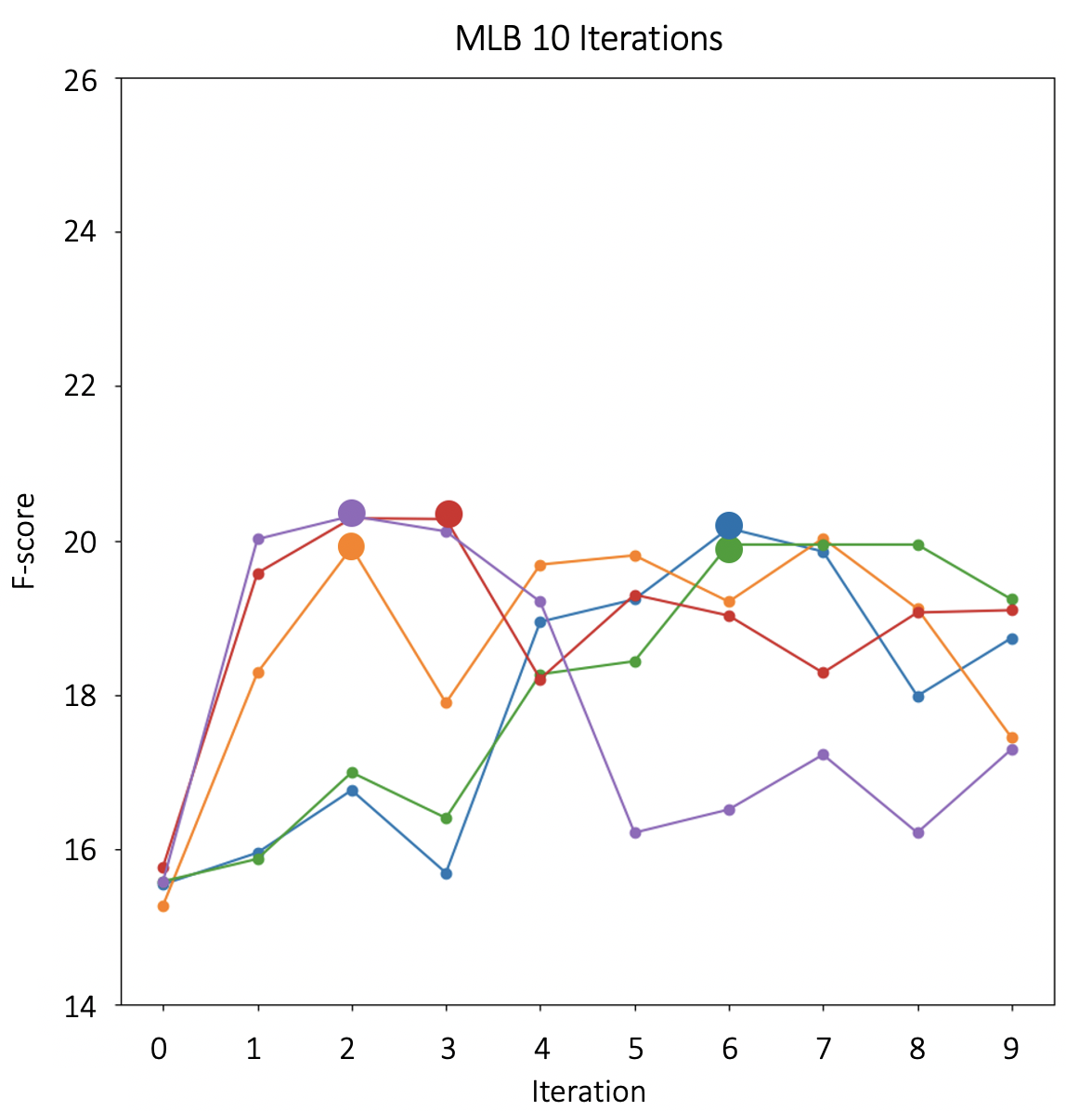}
        \caption{MLB.}
        \label{MLB10}
    \end{subfigure}
    \begin{subfigure}[b]{0.23\textwidth}
        \includegraphics[width=\textwidth]{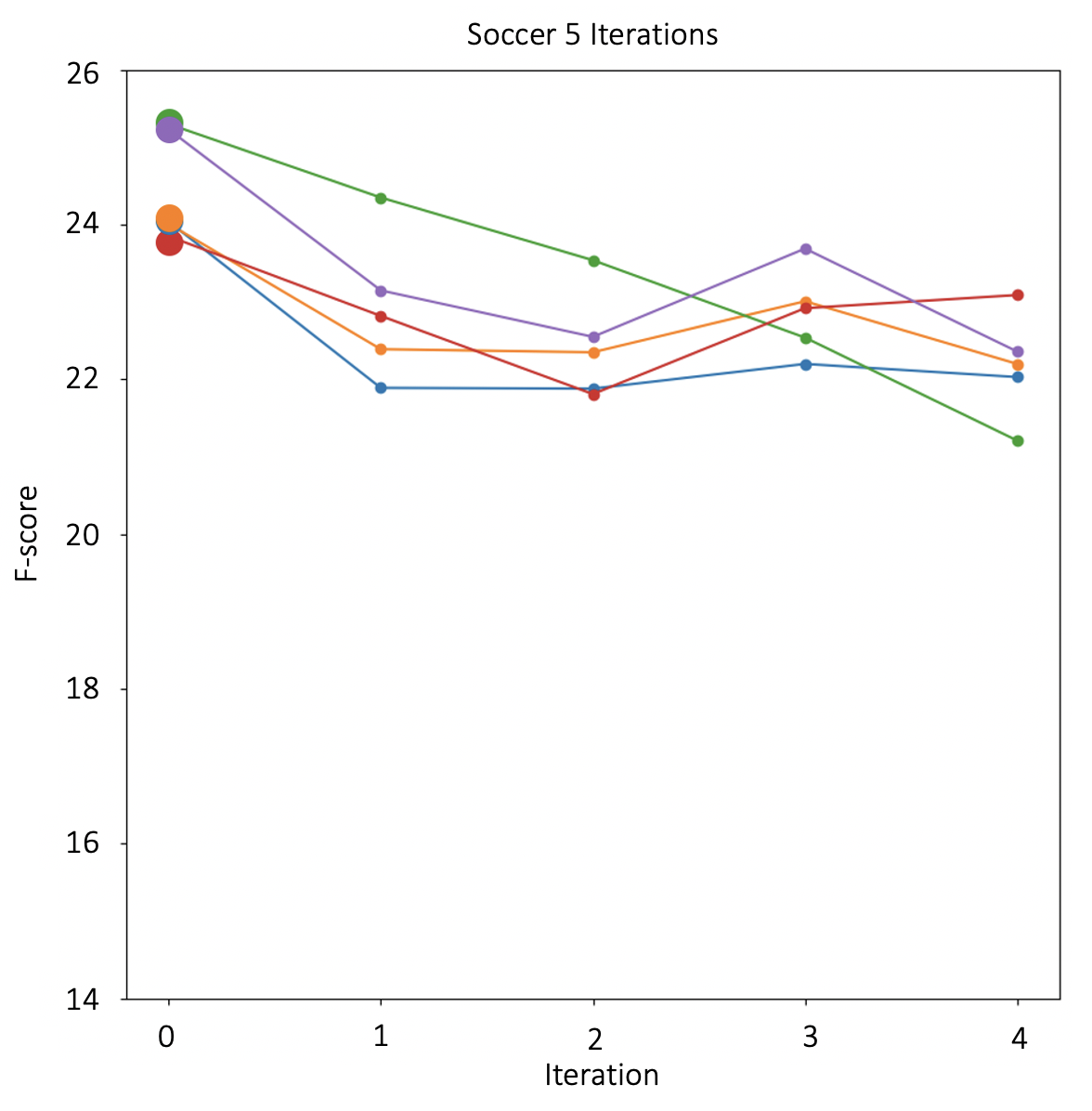}
        \caption{Soccer.}
        \label{Soccer5}
    \end{subfigure}
    \begin{subfigure}[b]{0.23\textwidth}
        \includegraphics[width=\textwidth]{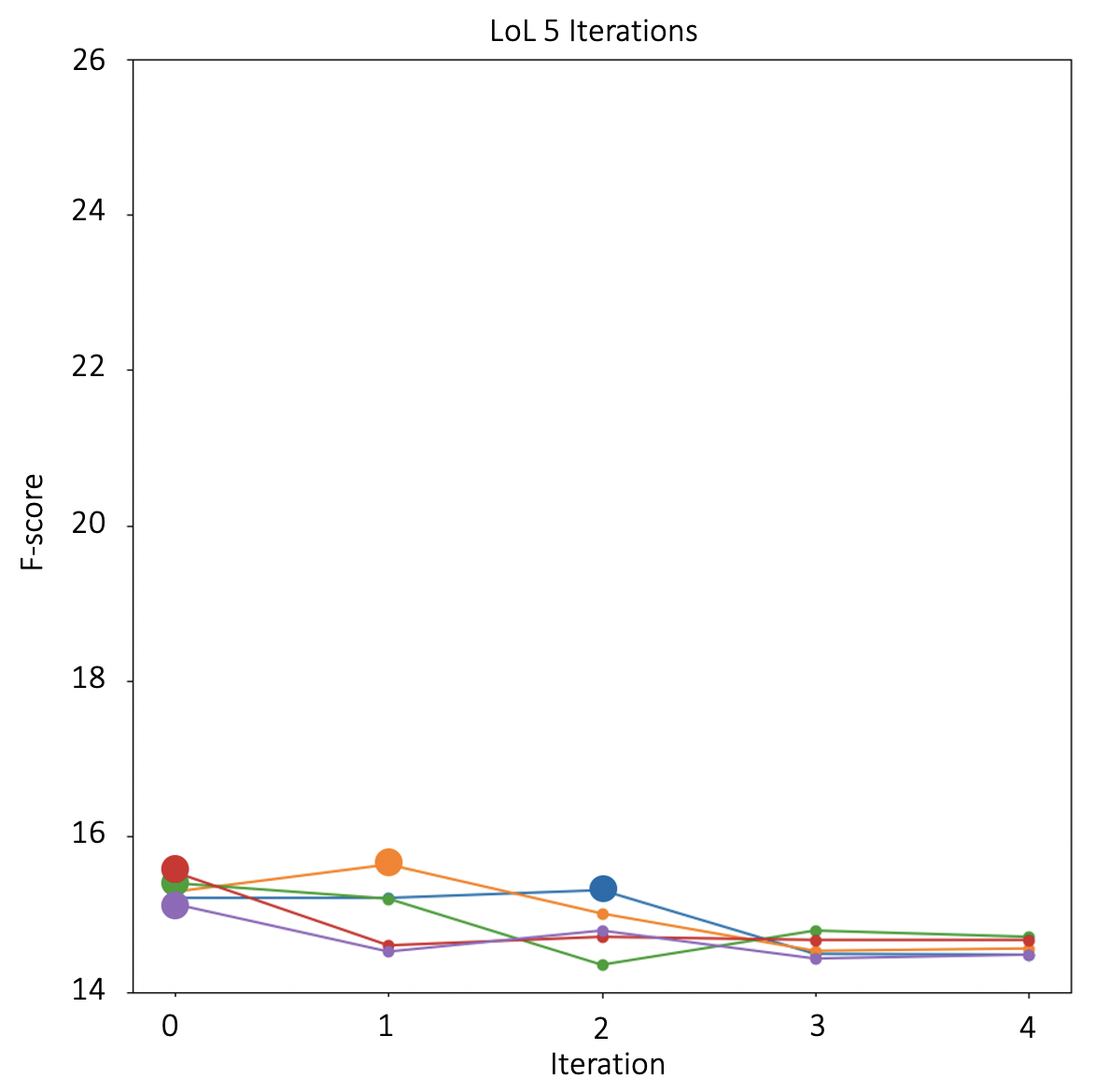}
        \caption{LoL.}
        \label{LoL5}
    \end{subfigure}
    \begin{subfigure}[b]{0.23\textwidth}
        \includegraphics[width=\textwidth]{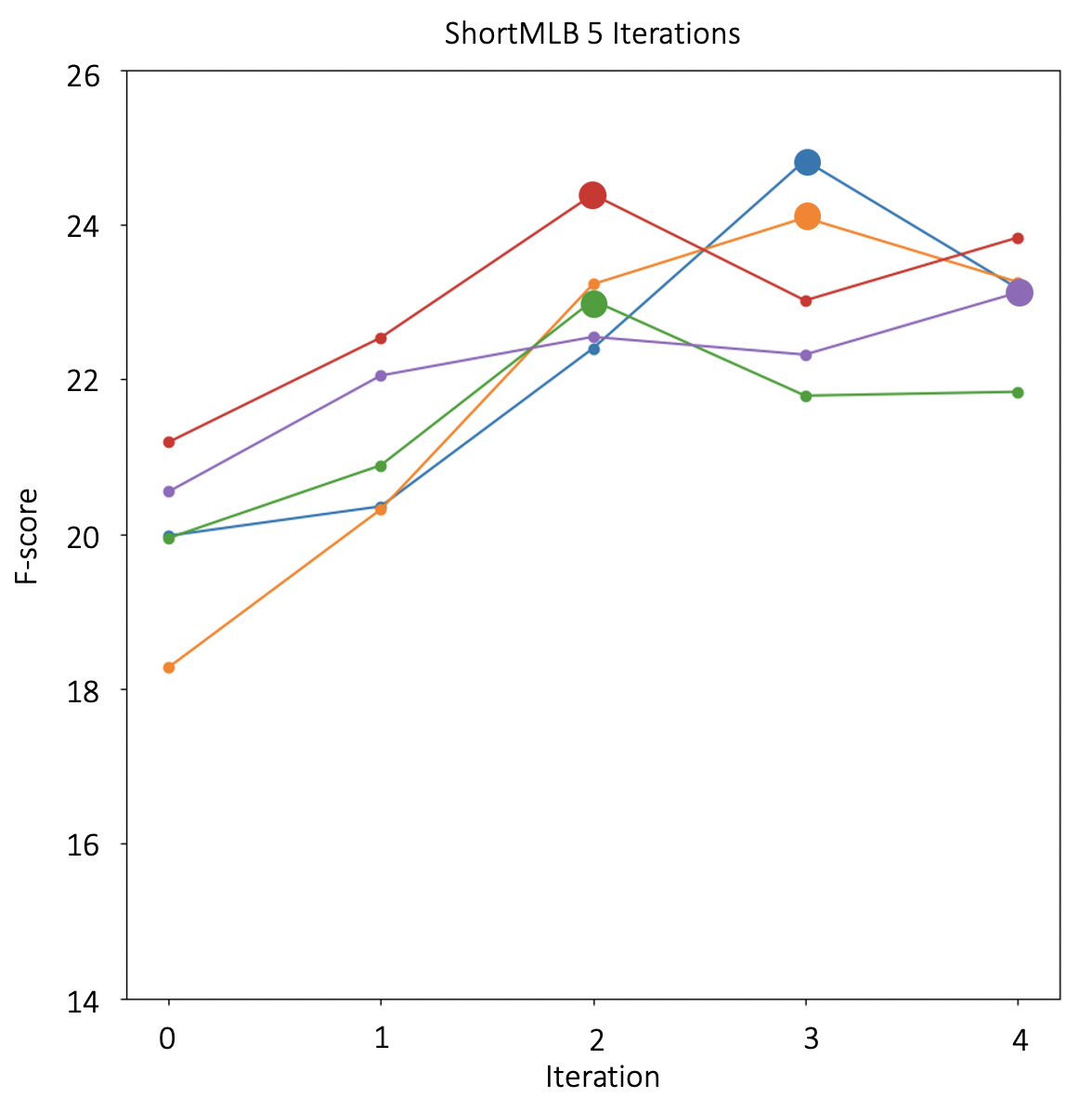}
        \caption{ShortMLB.}
        \label{ShortMLB5}
    \end{subfigure}
    \caption{SUM-SR$_{sepMa}$ performance (F-score ($\%$)) over multiple iterations on six datasets. In this figure, we display the result of five splits in one experiment for each dataset. We run SUM-SR$_{sepMa}$ for 10 iterations on SumMe (a) and MLB (c). We run SUM-SR$_{sepMa}$ for 5 iterations on other datasets.}
    \label{itersplit}
\end{figure}

\begin{figure}[tb]
    \centering
    \begin{subfigure}[b]{0.4\textwidth}
        \includegraphics[width=\textwidth]{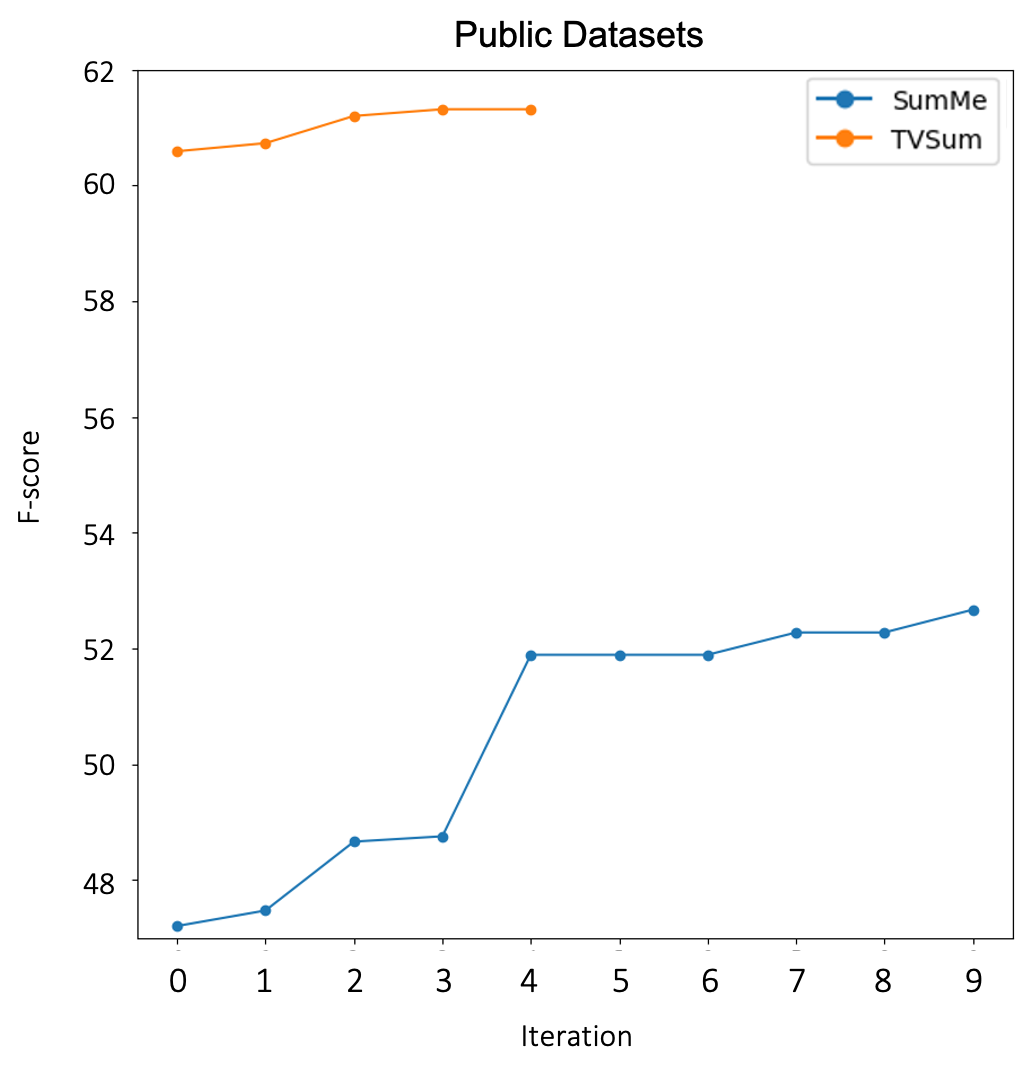}
        \caption{Public datasets.}
        \label{iterbestpub}
    \end{subfigure}
    \begin{subfigure}[b]{0.4\textwidth}
        \includegraphics[width=\textwidth]{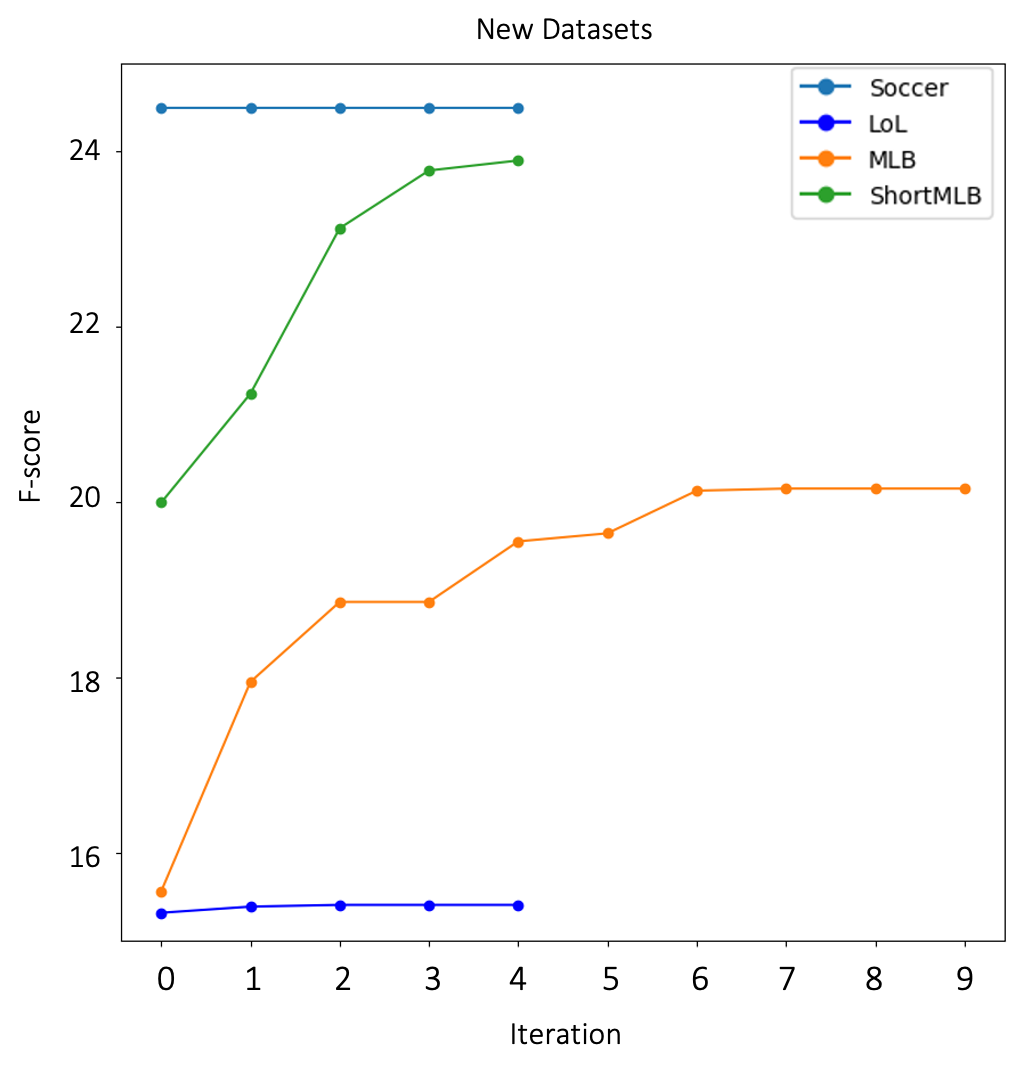}
        \caption{Datasets we created.}
        \label{iterbestnew}
    \end{subfigure}
    \caption{SUM-SR$_{sepMa}$'s best performance (F-score ($\%$)) over multiple iterations.}
    \label{iterbest}
\end{figure}

To explore more the impact of iterations, we plot the performance of SUM-SR$_{sepMa}$ on each dataset over multiple iterations in \cref{itersplit} and \cref{iterbest}. In each experiment, we run the model on five random splits on each dataset. In \cref{itersplit}, we plot the F-score of each split over multiple iterations. The results on SumMe and MLB for 10 iterations show that most splits reach their best performance in the first five iterations, with only a few splits reaching a slightly better performance after the fifth iteration. It is evident that five iterations are sufficient to improve model performance. Then we train the model on four other datasets for five iterations and record the remaining results in \cref{itersplit}. For both ShortMLB and MLB, the model performs better with the iterative training strategy of all splits. For SumMe and TVSum, most splits (four splits of SumMe and three splits of TVSum) have their best performance after iteration 0. For LoL and Soccer, the F-score drops after iteration 0 for most splits (three splits of LoL and all splits of Soccer). 

To better evaluate the contribution of the iterative training strategy, we select the highest F-score until the current iteration for each split and average them over five splits. The results are included in \cref{iterbest} as line charts. The more the iterative strategy improves the model performance, the more the curve increases. According to \cref{iterbest}, the iterative training strategy improves the model performance slightly on TVSum and remarkably on SumMe, MLB, and ShortMLB. For Soccer and LoL, the iterative training strategy has a minimal positive effect on the model performance, which explains the unsatisfactory performance of SUM-SR$_{5iter}$ on Soccer and LoL compared to other versions of the model. According to \cref{Soccer5}, SUM-SR$_{sepMa}$ always performs best in the first iteration on Soccer, suggesting that the model in iteration 2 to 5 overfits the training set. Although, our model selection method can select the best iteration at most of the time, a few failures to select the best iteration causing SUM-SR$_{5iter}$ performs slightly worse than SUM-SR$_{sepMa}$ on Soccer. In general, the iterative strategy improves the model performance on most datasets. Since the unsupervised model selection method can pick the best iteration most of the time, the model with iterative training performs well on all datasets.

\section{Examples}
\label{app:examples}
\begin{figure}[tb]
    \centering
    \begin{subfigure}[b]{0.45\textwidth}
        \includegraphics[width=\textwidth]{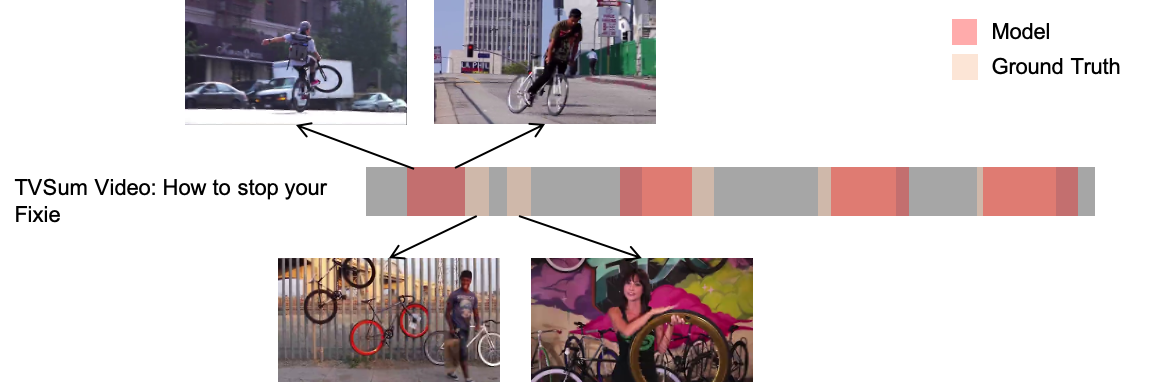}
        \caption{Selected shots demo on TVSum}
        \label{tvsum_demo}
    \end{subfigure}
    \begin{subfigure}[b]{0.45\textwidth}
        \includegraphics[width=\textwidth]{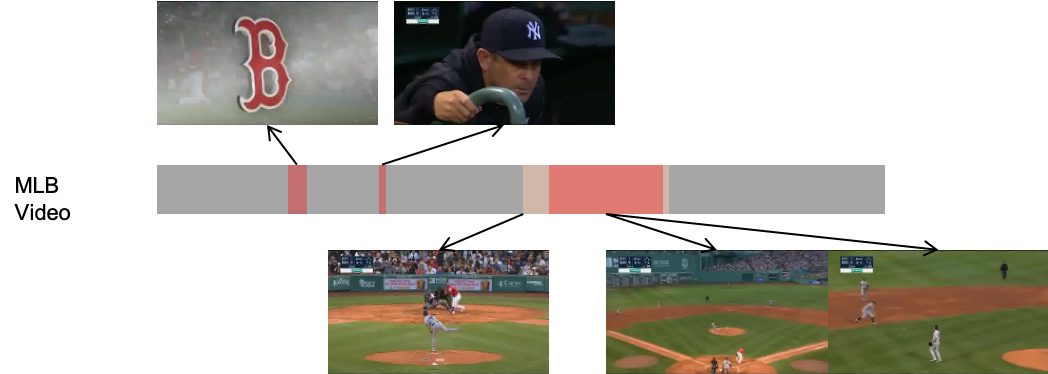}
        \caption{Selected shots demo on MLB}
        \label{mlb_demo}
    \end{subfigure}
    \caption{Sample demonstrations of model generated summaries and user-generated summaries.}
    \label{summary_demo}
\end{figure}

To examine the discrepancy between the summary generated by the model and the reference summary, we analyze two test videos selected from TVSum and MLB. The visual representations of these cases are illustrated in \cref{summary_demo}. In \cref{tvsum_demo}, the model's choice encompasses two consecutive shots featuring a person riding a bicycle haphazardly, whereas the ground truth summary comprises of two other shots depicting a person positioned on the curb and a host demonstrating the wheel. In \cref{mlb_demo}, the model-generated summary includes a brief clip of the moving team logo "B", which is absent in the ground truth summary. Additionally, it selects a shot of the camera tracing the ball and infielders fielding a ground ball, yet it omits capturing the hit at the previous at bat. Based on these observations, the model prefers video shots with moving objects and a dynamic camera view. Also, the model prefers video shots that are distinct from the overall content of the video. In \cref{mlb_demo}, the model picks a shot of the team logo in motion (distinct property) and a shot of a manager in a dugout (also distinct due to his dark uniform), which have no similar content in the rest of the video.
\end{document}